# Monocular depth perception enhancement based on joint shading/contrast model and motion parallax (JSM)

*Seungchul Ryu, Hyunjin Yoo, and Tara Akhavan*
**Faurecia Irystec Inc., Montreal, QC, Canada**

## Abstract

***Stereoscopic 3D displays adopt a binocular depth cue to provide depth perception. However, users should be equipped with expensive special devices to appreciate depth perception based on the binocular depth cues. Also, visual fatigue induced by the stereoscopic display is still a challenging open problem. In order to overcome this limitation, this paper proposes a novel framework, JSM, to enhance monocular depth perception, significantly improving both depth volume perception and depth range perception. The proposed framework can not only provide an enhanced depth perception on any conventional 2D display devices, but also it can be applicable to the 3D display devices since it is complimentary for binocular depth cues. The qualitative evaluation, ablation study, and subjective user evaluation proved the advantages and practicability of the proposed framework.***

## 1. Introduction

Depth perception is the key visual ability for humans to perceive the world in 3D, especially the distance between objects. However, the traditional display technologies can only display images with limited depth perception. In order to overcome this limitation, in recent decades, many methods and devices have been developed to provide enhanced depth perception. Due to the significant advantages of allowing to perceive more realistic image contents, depth perception enhancement has attracted much attention in display and image processing applications [1].

Depth impression is derived based on two cues: binocular and monocular depth cues. The majority of researches [2, 3] were focused on the binocular depth cue to provide depth perception due to its extreme effects. However, users should be equipped with expensive special devices such as stereoscopic display devices to appreciate depth perception based on the binocular depth cues. Also, visual fatigue induced by the stereoscopic display is still a challenging open problem [4, 5]. Thus, several works [6, 7] studied to enhance monocular depth cues as an alternative approach, which does not require any specialized 3D display and does not induce visual fatigue.

The monocular depth cues help us perceive depth when viewing a 2D image, including linear perspective, occlusion, aerial perspective, contrast, motion parallax, and shading cues [8, 9, 10, 11]. Depth perception cues determine the following two types of perception: depth volume perception and depth range perception. Depth volume perception refers to the perceived volume of each object in an image, and depth range refers to the perceived distance within the entire image.

One approach direction to enhance depth perception was an image enhancement from a single image by either requiring or estimating the depth information [12, 13, 14]. These methods directly rely on the depth information of the scene for enhancement, which is assumed as given, either 3D models, depth maps, or z-buffer values.

Another direction was to enhance depth perception based on monocular depth cues. Depth perception was induced by enhancing area, texture, and luminance contrast [15, 16]. The contrast-based depth induction could explain the depth-from-HDR effect [17]. Jung et al. [6] proposed to enhance the depth perception of single-view images based on the linear perspective, aerial perspective, focus, and shadow effects. Hel-Or et al. [7] recently proposed a tone mapping operator that controls the shading component of an image to enhance depth perception. However, those methods were limited in: modeling monocular depth cues separately and only considering one of depth volume perception or depth range perception.

This paper proposes a novel framework to enhance monocular depth perception, significantly improving both depth volume perception and depth range perception via joint modeling of shading and contrast. Further, two types of motion parallax are included in the framework. The main contributions of this paper are summarized as follows.

- This paper proposes a general framework, JSM, to enhance monocular depth perception in terms of both depth volume and range for the first time.
- This paper proposes an effective and efficient joint model of shading and contrast that can provide adjustable depth perception.
- The proposed framework can not only provide an enhanced depth perception on any conventional 2D display devices, but also it can be applicable to the 3D display devices since it is complimentary for binocular depth cues.

The remaining of this paper is organized as follows. Section 2 describes the proposed monocular depth perception enhancement based on the joint shading/contrast model and motion parallax (JSM) in detail. Experimental results are given in Section 3. Lastly, Section 4 concludes this paper with some future works.

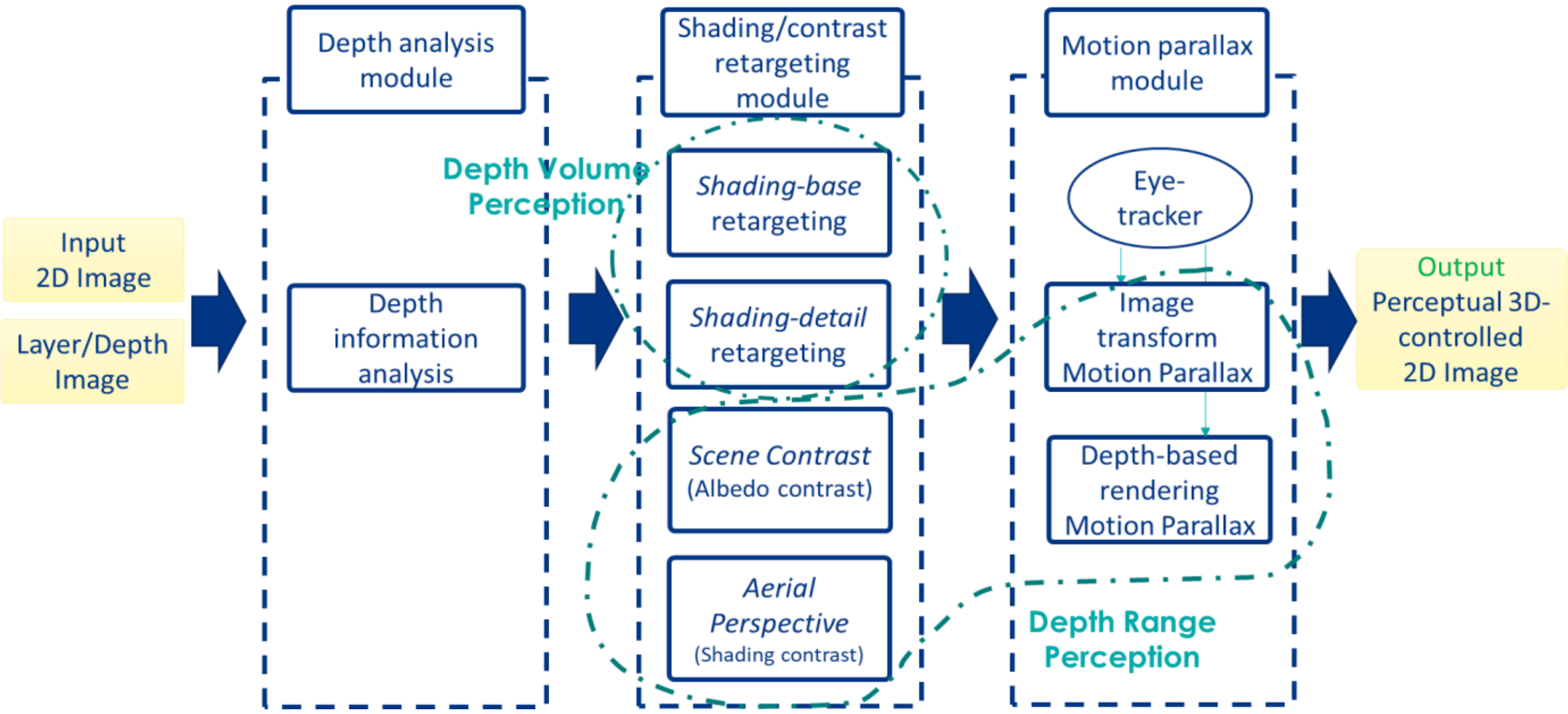

**Figure 1.** The diagram of the proposed JSM framework

## 2. The proposed JSM framework

The proposed JSM framework consists of the depth analysis module, shading/contrast retargeting module, and motion parallax module, as depicted in Fig. 1. The depth analysis module estimates the pixel-wise depth information to determine foreground and background. The shading/contrast retargeting and motion parallax modules are controlled according to the depth profiles obtained from the depth analysis module. That is, the amount of the shading cue, contrast cues, and motion parallax are based on the depth profile. Here, two types of depth profiles are implemented: two-layered depth and continuous depth.

The shading/contrast retargeting module jointly models the shading, local texture, and global contrast cues and adjusts each element to enhance the depth perception. In the joint model, the shading and local texture cues contribute to increasing the depth volume perception, while the global contrast cue contributes to increasing the depth range perception. In the joint model, the luminance of an input color image $\mathbf{I}_{\mathrm{y}}$ is modeled by the multiplication of albedo $\mathbf{A}_{\mathrm{y}}$ and shading $\mathbf{S}$, as follows:

$$\begin{aligned} \mathbf{I}_{\mathrm{y}} &= \mathbf{A}_{\mathrm{y}} \cdot \mathbf{S} \\ &= \mathbf{A}_{\mathrm{y}} \cdot (\mathbf{S}_{\mathrm{B}} + \mathbf{S}_{\mathrm{D}}) \end{aligned} \quad (1)$$

where $\mathbf{S}_{\mathrm{B}}$ and $\mathbf{S}_{\mathrm{D}}$ represent the base layer and detail layer of shading. $\mathbf{A}_{\mathrm{y}}$ is computed by the guided filter-based albedo extraction algorithm, which is formulated as follows:

$$\mathbf{A}_{\mathrm{y}} = <\mathbf{A}_{\mathrm{rgb}}> \quad (2)$$

$$\mathbf{A}_{\mathrm{rgb}} = f_{\mathbf{A}}(\mathbf{I}_{\mathrm{rgb}}) \quad (3)$$

where $\mathbf{I}_{\mathrm{rgb}}$ is the input color image, and $f_{\mathbf{A}}$ indicates the guided filter. The operation <X> indicates the selection of the luminance channel of X.

The base-shading information $\mathbf{S}_{\mathrm{B}}$ is retargeted with a non-linear tone-mapping operator to control the shading depth volume cue, which is formulated as follows:

$$\mathbf{S}'_{\mathrm{B}} = g_n(\mathbf{S}_{\mathrm{B}}) \quad (4)$$

where $\mathbf{S}'_{\mathrm{B}}$ is the retargeted base-shading, and $g_n$ is a non-linear tone-mapping operator (here, the truncated power function is used). Then, the detail-shading information $\mathbf{S}_{\mathrm{D}}$ is enhanced with a linear boosting operator to control the shading depth volume cue, which is defined as follows:

$$\mathbf{S}'_{\mathrm{D}} = g_l(\mathbf{S}_{\mathrm{D}}) \quad (5)$$

where $\mathbf{S}'_{\mathrm{D}}$ is the boosted detail-shading, and $g_l$ is a linear boosting operator.

Further, the depth range perception is increased by controlling the contrast cues in the entire image, where three types of contrast are adopted: albedo contrast, shading contrast, and texture contrast. Albedo contrast refers to the dynamic range within the entire image, which is increased to give a higher depth range perception. Shading and texture contrast respectively refer to the shading difference and the details difference between

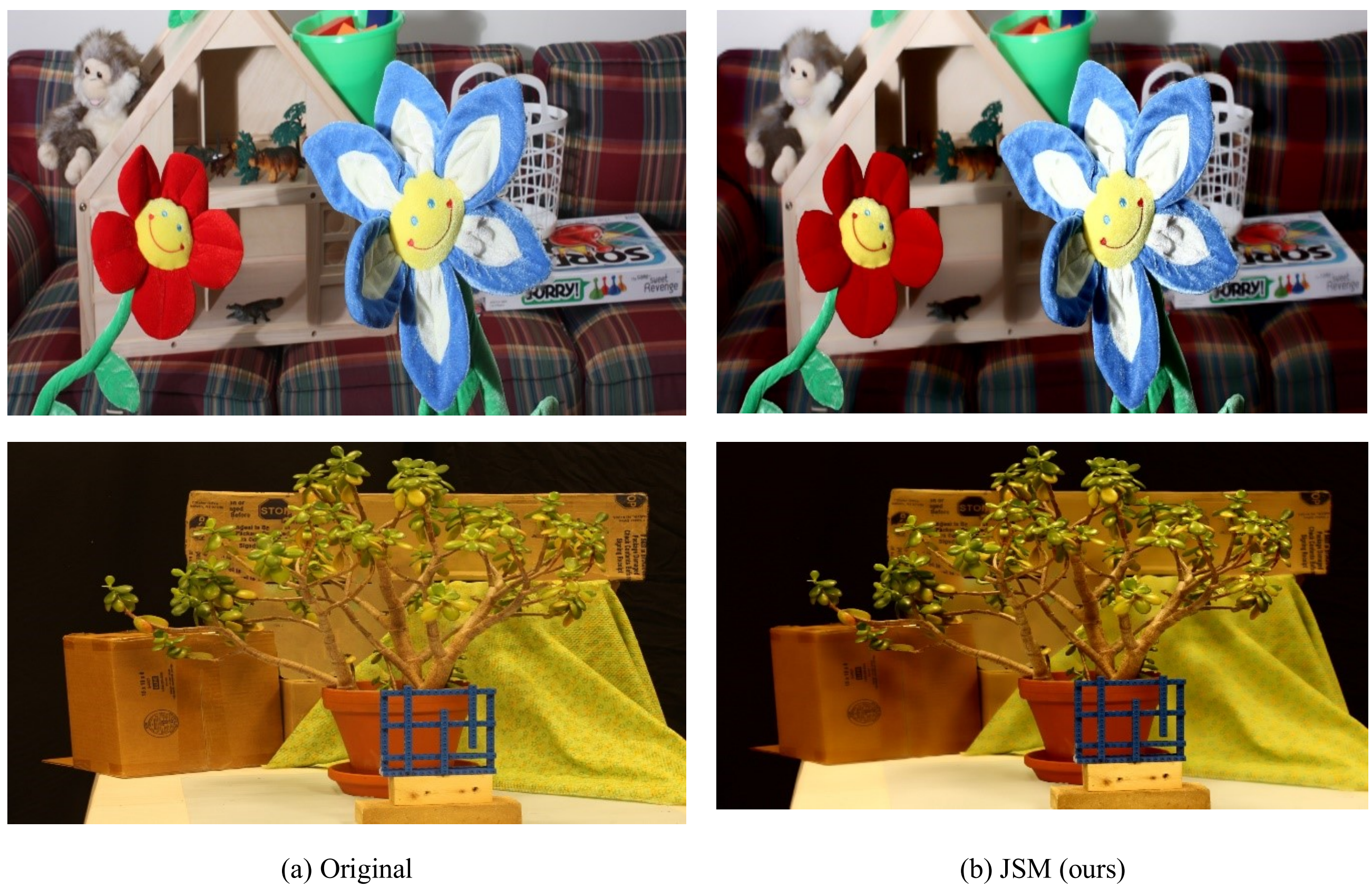

(a) Original (b) JSM (ours)

**Figure 2.** Qualitative evaluation of the JSM model.

foreground and background. Due to the aerial perspective, closer objects appear more shading and textures while distant objects are less shading and more out-focused. In order to manipulate the shading contrast and texture contrast, in the proposed joint model, base-shading and detail-shading are emphasized for the closer objects while suppressing shading information for the far objects according to the depth profile. Besides, global luminance contrast is controlled in albedo image as follows:

$$\mathbf{A}'_{\mathrm{y}} = g_t(\mathbf{A}_{\mathrm{y}}) \tag{6}$$

where $\mathbf{A}'_{\mathrm{y}}$ is the retargeted albedo image via the tone-mapping operator $g_t$.

The processed albedo, base-shading, and detail-shading are composed to obtain the depth perception enhanced luminance $\mathbf{I}'_{\mathrm{y}}$ as follows:

$$\mathbf{I}'_{\mathrm{y}} = \mathbf{A}'_{\mathrm{y}} \cdot (\mathbf{S}'_{\mathrm{B}} + \mathbf{S}'_{\mathrm{D}}) \tag{7}$$

The shading/contrast retargeting module is followed by the motion parallax module that increases the depth range perception by providing moving objects according to their relative depth. For example, the closer objects are moving faster than the far objects. In the proposed framework, this movement is implemented as being relative to the viewer's head movement.

## 3. Experimental Results

In order to evaluate the proposed framework, we conducted the qualitative evaluation, ablation study, and subjective user evaluation on the natural images and automotive cluster images. The images were collected from the public Middlebury [18] dataset and our own generated dataset. All the images were resized to 1920x1080 and processed on Intel(R) Core™ i7-10750H 2.60GHz. Fig. 2 shows the qualitative evaluation, indicating that the proposed framework can enhance depth perception in terms of depth volume and range. The ablation study analyzed and proved the contribution of each element in the proposed framework. As shown in Fig. 3, each element enhances the depth perception in terms of either depth volume or depth range.

Further, subjective user evaluation was conducted to compare the proposed framework with the autostereoscopic 3D display (3D Global Solutions) and the 2D original image (Table 1). The depth perception, visual fatigue, and overall 3D experience were evaluated by 18 participants whose ages ranged from 20-50, and the evaluation scores were scaled into the range of 0~10. The evaluation results show that the proposed framework can achieve comparable depth perception but very low visual fatigue than the autostereoscopic 3D display.

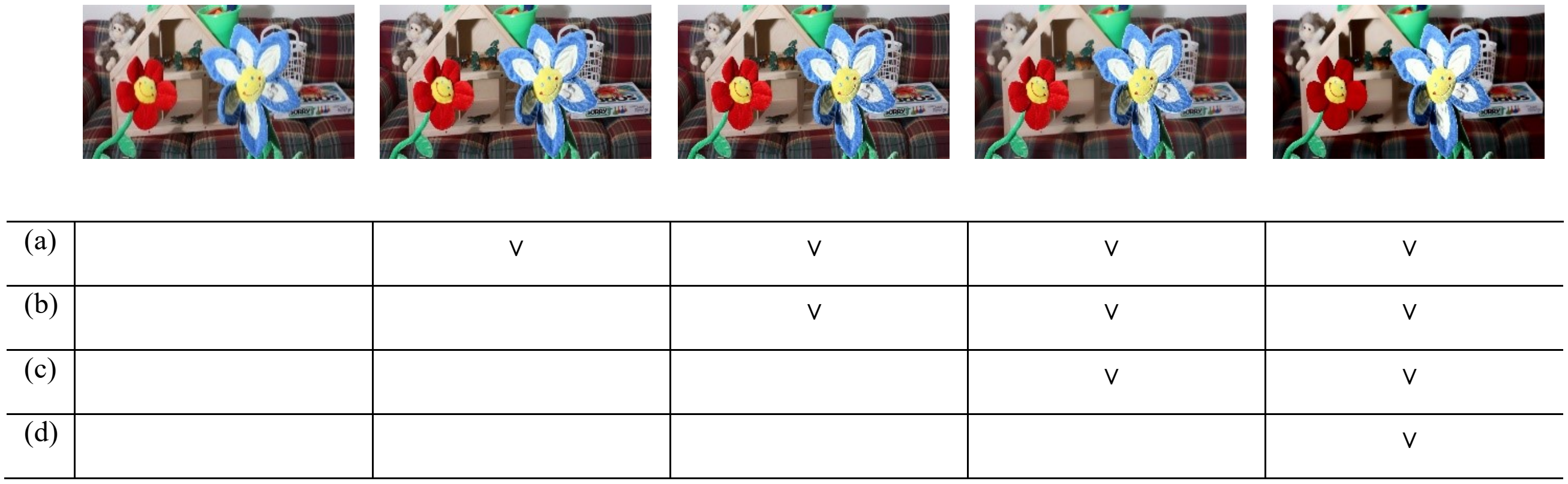

| | | | | | |
|---|---|---|---|---|---|
| (a) | | ∨ | ∨ | ∨ | ∨ |
| (b) | | | ∨ | ∨ | ∨ |
| (c) | | | | ∨ | ∨ |
| (d) | | | | | ∨ |

**Figure 3.** Ablation study of the JSM model. ∨ indicates that the sub-module is enabled (a) base-shading, (b) detail-shading, (c) shading contrast, and (d) albedo contrast. That is, the left-most and right-most represent the original image and the JSM result.

**Table 1.** Subjective user evaluation for comparison of the 2D display, 3D display, and JSM (ours).

| | Depth Perception | Visual Fatigue | Overall 3D Experience |
|---|---|---|---|
| 2D Display | 2.3 | 0.5 | 2.0 |
| JSM on 2D Display | 5.5 | 1.0 | 6.0 |
| 3D Display | 8.2 | 8.5 | 6.7 |